\documentclass[10pt,twocolumn,letterpaper]{article}

\usepackage[pagenumbers]{cvpr} 

\usepackage{comment}
\usepackage{amsmath,amssymb} 
\usepackage{color}
\usepackage{graphicx}
\usepackage{hyperref}       
\usepackage{url}            
\usepackage{booktabs}       
\usepackage{amsfonts}       
\usepackage{nicefrac}       
\usepackage{microtype}      
\usepackage{xcolor}         
\usepackage{wrapfig}
\usepackage{listings}
\usepackage[frozencache,cachedir=.]{minted}
\usepackage{algpseudocode}
\usepackage{algorithm}
\usepackage{flexisym}
\usepackage{multirow}
\usepackage{xspace}
\usepackage{makecell}

\usepackage[capitalize]{cleveref}
\crefname{section}{Sec.}{Secs.}
\Crefname{section}{Section}{Sections}
\Crefname{table}{Table}{Tables}
\crefname{table}{Tab.}{Tabs.}

\begin{document}

\newcommand{\citep}{\cite}
\newcommand*{\projectname}{FedWeight\xspace}
\title{Federated Remote Physiological Measurement with Imperfect Data}

\author{%
  Xin Liu\textsuperscript{1},
  Mingchuan Zhang\textsuperscript{3},
  Ziheng Jiang\textsuperscript{1},
  Shwetak Patel\textsuperscript{1},
  Daniel McDuff\textsuperscript{2}\\
  Paul G. Allen School of Computer Science \& Engineering, University of Washington, Seattle, USA\textsuperscript{1}\\
  Microsoft Research, Redmond, USA\textsuperscript{2}\\ 
  School of Computer Science,
  Fudan University, Shanghai, China\textsuperscript{3}
 \\
 \{xliu0, ziheng, shwetak\}@cs.washington.edu,\\ damcduff@microsoft.com, mczhang18@fudan.edu.cn
 }

\maketitle

\begin{abstract}
The growing need for technology that supports remote healthcare is being acutely highlighted by an aging population and the COVID-19 pandemic. In health-related machine learning applications the ability to learn predictive models without data leaving a private device is attractive, especially when these data might contain features (e.g., photographs or videos of the body) that make identifying a subject trivial and/or the training data volume is large (e.g., uncompressed video). Camera-based remote physiological sensing facilitates scalable and low-cost measurement, but is a prime example of a task that involves analysing high bit-rate videos containing identifiable images and sensitive health information.
Federated learning enables privacy-preserving decentralized training which has several properties beneficial for camera-based sensing. We develop the first mobile federated learning camera-based sensing system and show that it can perform competitively with traditional state-of-the-art supervised approaches. However, in the presence of corrupted data (e.g., video or label noise) from a few devices the performance of weight averaging quickly degrades. To address this, we leverage knowledge about the expected noise profile within the video to intelligently adjust how the model weights are averaged on the server. Our results show that this significantly improves upon the robustness of models even when the signal-to-noise ratio is low.

\end{abstract}
\section{Introduction}
Federated learning (FL) enables distributed devices (e.g., cellphones) to collaboratively learn models without data leaving each device~\cite{mcmahan2017communication, konevcny2016federated}. While creating traditional machine learning systems involves uploading raw data and labels to a centralized location for training, FL can avoid this. A core premise is that a model trained from aggregated decentralized data can be more effective than training with the data that any one device has access to on its own. More specifically, federated learning leverages locally-computed updates (weights) from a large number of single devices to create a robust aggregated model that can then be shared. To summarize, federated learning has several useful properties, the ability to: 1) preserve privacy more easily by only sharing model weights instead of raw data and labels, 2) increase the diversity and generalizability of a model by aggregating a diverse population's data, 3) reduce the bandwidth and storage resources required when uploading raw data to a centralized server. 

\begin{figure}[t]
  \centering
  \includegraphics[width=0.9\columnwidth]{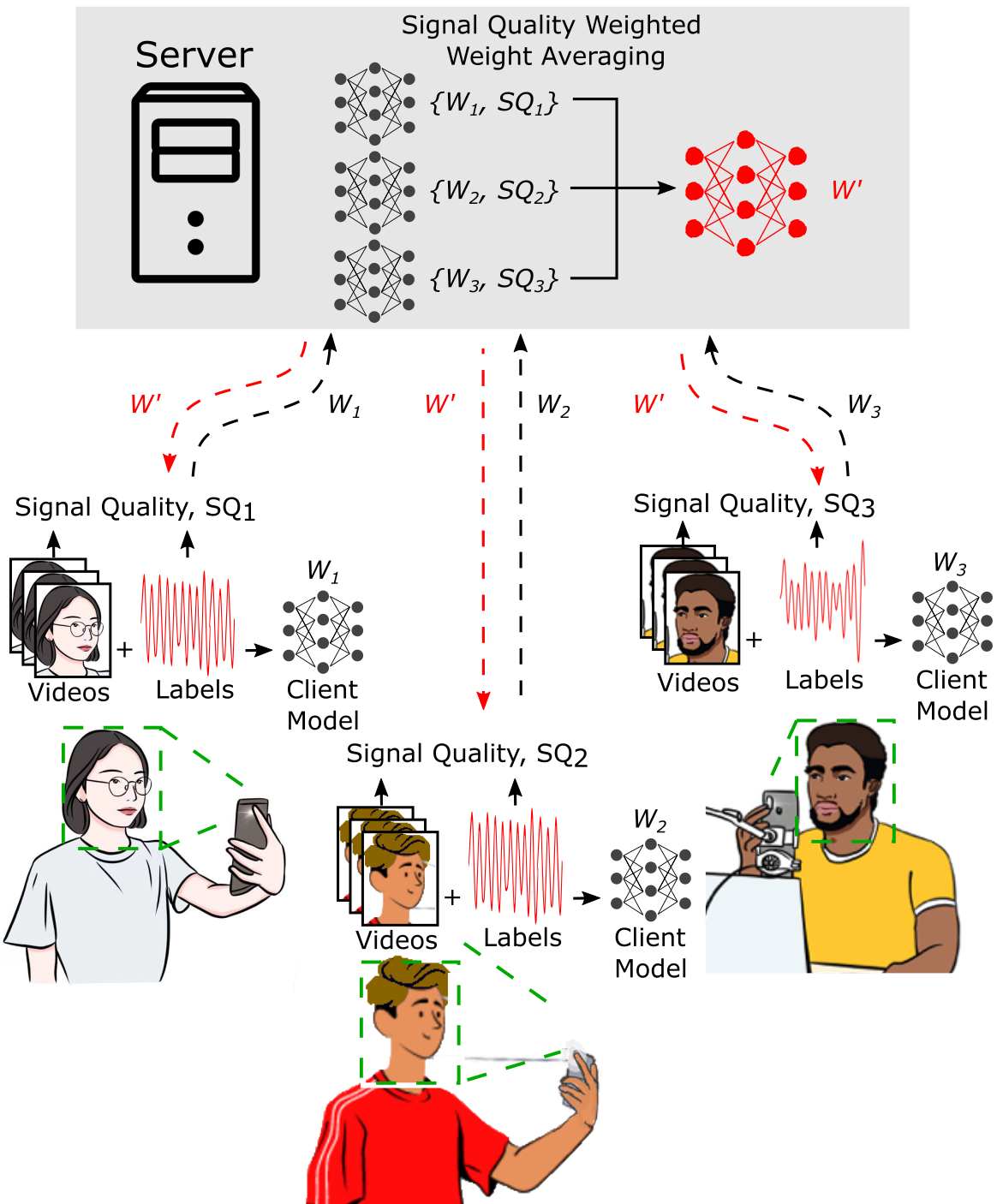}
  \caption{We present a privacy preserving federated system for on-device, camera-based physiological sensing. We propose a novel weight averaging approach that significantly improves on model robustness in the presence of noisy videos and labels. $W_{N}$ represents the weights from each client, SQ$_{N}$ represents the signal quality score either for the video, labels or both, and $W'$ represents the server weights after weight averaging.}
   \label{fig:onecol}
\end{figure}

The benefits of FL are particularly attractive in applications in which models rely on sensitive data that are also personally identifiable. This is very true in contexts that involve biometric, physiological and health data. The growing need for technology that supports remote healthcare has been acutely highlighted by the COVID-19 pandemic~\cite{xu2020pathological, zheng2020covid}. One such technology that can support remote care is low-cost, on-device, camera-based vital sign measurement~\cite{verkruysse2008remote,poh2010advancements,li2014remote,wang2016algorithmic,chen2018deepphys,liu2020multi}. These systems use ubiquitously available webcams and smartphone cameras to measure important physiological vital signs such as the cardiac pulse~\cite{poh2010non}, breathing rate~\cite{poh2010advancements} and blood oxygen saturation~\cite{tarassenko2014non} of a patient without the data leaving the device. The methods rely on capturing subtle variations in light reflected from the body that capture volumetric changes in blood (the photoplethysmogram/PPG) and mechanical motions resulting from cardiac and respiratory function (e.g., the ballistocardiogram/BCG)~\cite{mcduff2021camera}. Democratizing (or scaling) camera-based physiological sensing in this way has much potential. For example, to help in screening for atrial fibrillation and other forms of arrhythmia~\cite{chan2016diagnostic} which are predictors of stroke risk.

Video recordings that contain the necessary fidelity to capture physiological changes contain \emph{both} private health data \emph{and} personally identifiable information. The physiological signals themselves have personally identifiable features~\cite{hernandez2015bioinsights} and the video frames may also contain visually recognizable body parts (e.g., the face). Furthermore, to effectively measure the very subtle changes in the body associated with these physiological processes, the videos should not be compressed too heavily as motion-compression algorithms typically remove the signals of interest~\cite{mcduff2017impact}. As such, the recordings contain sensitive data and are often large; therefore, they ideally would not be transferred or stored in great volumes in the cloud.

When building models for measuring physiological vital signs, it is critical that the learned representations are not corrupted because of ``bad'' data (either features or labels) from a few devices. However in the context of FL where the server does not have access to the data itself, how do we ensure that that this does not happen? 
Ideally, during weight aggregation it would be possible to adapt to, or \textit{exploit}, client weights that were derived from cleaner rather than noisier data. At the same time, we do not want to completely ignore weights from a given client as every client will have access to data from a subject that was not ``seen'' by other clients and generally we would want a model to \textit{explore} and maximize the diversity of our observations.

As shown in Fig.~\ref{fig:onecol}, in our scenario we have individuals collecting video on their own mobile devices alongside reference sensor measurements for training (as in~\cite{liu2021metaphys}). In this case, there could be different levels of video noise resulting from camera sensor quality and automatic gain calibration. There could also be noise in the reference label, for example if a person was moving during the calibration period or did not attach the reference sensor correctly. Fortunately, both video and the physiological signals of interest (i.e., the PPG signal) have been studied extensively. We have strong statistical priors about the nature of these signals. In this work, to demonstrate our approach clearly we perform experiments assuming knowledge about the signal-to-noise ratios in the videos and labels. However, we could equally leverage domain knowledge to automatically calculate weight contributions from different devices. Our method does not discard the weights from clients with noisy data, but rather includes all weights while accounting for signal quality.

The contributions of this paper are: 1) to introduce the first federated camera-based remote physiological measurement system, 2) to show that this system can match the performance of a traditional supervised learning approach, 3) to introduce a critical averaging approach that accounts for the signal quality and diversity of samples. 4) to provide an on-device mobile training and inference implementation. Our code, models, and video figures are provided in the supplementary materials.

\section{Related Work}

\textbf{Federated Learning in Healthcare.}
Federated learning enables training machine learning models from a set of distributed remote devices (e.g., mobile devices) while storing data only on the individual clients. Early work established optimization principals on how to perform non-convex optimization on distributed client's model weights~\cite{mcmahan2017communication}. Due to federated learning's unique characteristics in protecting privacy, it has been used and studied in healthcare applications. The volume of training data in healthcare applications is often smaller than in many traditional machine learning tasks. Therefore, aggregating as much data as possible from decentralized clients' could help boost the performance of machine learning applications in healthcare while not leaking sensitive information or violating HIPAA guidelines \cite{yang2019federated, rieke2020future}. Brisimi et al. \cite{brisimi2018federated} proposed to use federated learning to train a supervised classification model for cardiac events. More specifically, they develop a federated learning based framework to enable multiple data holders (i.e., hospitals) to collaborate and converge to a centralized model. More recently, \cite{chen2020fedhealth} proposed a framework that leveraged federated learning to perform transfer learning for wearable sensors called FedHealth. In this framework, when the clients receive the updated model weights from the server all the layers in the neural network are frozen except for the last two fully connected dense layers. They claim that fine-tuning the last two layers on the client side can help build personalized models for each user or organization. FedHealth was evaluated on a Parkinson’s disease dataset. The application of federated learning in COVID-19 has also been investigated. Qayyum et al.~\cite{qayyum2021collaborative} explored the use of federated learning in automatic diagnosis of COVID-19. They demonstrated improvements on results of X-ray and Ultrasound datasets after using federated learning. In the field of physiological measurement, Brophy et al. \cite{brophy2021estimation} investigated the use of federated learning and generative adversarial networks to estimate continuous blood pressure from the PPG signal. This work is quite distinct from ours as it uses contact sensor based PPG measurements while our work is focused on deriving the PPG signal and heart rate from facial videos. 

\textbf{Machine Learning in Remote Physiological Measurement.}
Remote physiological measurement or camera based physiological measurement is an emerging field. Early research established signal processing based methods for extracting physiological signals (in particular the cardiac pulse) from light reflection capture by the camera ~\cite{takano2007heart, verkruysse2008remote, poh2010advancements, de2013robust, tulyakov2016self, li2014remote, wang2016algorithmic}. For example, Independent Component Analysis (ICA) was proposed to demix RGB channel information to recover a source containing the blood volume pulse (BVP)~\cite{poh2010advancements}. Wang et. al further extended this by calculating a projection plane orthogonal to the skin-tone based on physical principles \cite{wang2016algorithmic}. Similar to many other vision tasks, deep learning has also helped boost the performance of remote physiological sensing, making models more robust to sources of noise seen in real-world applications including head motions and ambient lighting changes. A two-branch convolutional attention neural network was first proposed~\cite{chen2018deepphys}. To model spatial and temporal information from the videos simultaneously, a 3D convolutional neural network was presented to further improve performance~\cite{yu2019remote}. More recently, an on-device Temporal Shift Convolutional Attention Network (TS-CAN) was proposed to address the gap between efficiency and accuracy~\cite{liu2020multi}. TS-CAN achieved state-of-art accuracy while dramatically reducing the computational cost and enabling real-time demonstrations on an embedded system at a high frame rate. Researchers have also investigated meta learning as a way to perform few-shot adaption for personalizing camera-based physiological sensing models~\cite{lee2020meta,liu2021metaphys}.

\section{Method}

Traditional supervised learning approaches to camera-based physiological sensing have been trained on large-scale centralized video datasets and physiological labels~\cite{chen2018deepphys,yu2019remote,liu2021metaphys}. There are several drawbacks to this. First, the data are highly identifiable containing appearance (e.g., faces) and physiological information. Second, these data consume considerable data storage resources (data for each subject often excess 1GB). For these reasons it would be desirable to have a solution that only involves analyzing videos on the client (so that videos need not be shared) and ideally in distributed manner.
In this paper, we explore the use of federated learning in camera-based video-based physiological measurement.
We leverage domain knowledge about the expected noise profile within our data to intelligently dynamically adjust how the model weights are averaged on the server. Our results empirically show that approach creates a more accurate physiological estimation model.
\begin{algorithm}[ht]
\label{alg: mobilephys_alg_train}
\caption{\projectname: Federated Remote Physiological Measurement with Signal Quality Weighting}
\begin{algorithmic}[1]
\Require $S$: Subject-wise video data
\State \textbf{Server Update: } with an initialization $W_0$ 
\For {each round $t = 1, 2, 3 ...$}
\State $S_t \leftarrow$  random select a set of clients
\For{each client k in $S_t$}
\State $\omega^k_{t}, b^k_{t}, \sigma_k = ClientUpdate(k, W_t)$
\EndFor
\State  $W_{t} = \frac{\sigma_k}{\sum{\sigma_{k}}} \cdot (\omega^k_t + b^k_t)$
\EndFor
\State \textbf{Client Update: (k, $\theta$)}
\For{each batch B in }
\State $\omega^k_{t}, b^k_{t} \leftarrow \theta - \beta \nabla_{\theta}$ $ \mathcal L(f(\theta))$
\State $\sigma_k$ $\leftarrow$ assessing signal quality of client k based on noisy levels 
\EndFor
\end{algorithmic}
\end{algorithm}

\textbf{Federated Learning based Video-based Physiological Measurement.}
FL is a decentralized training schema where clients (i.e., smartphones) perform local training and upload trained model weights to a centralized server (e.g., the cloud). This training mechanism minimizes the risks associated with leaking identifiable or sensitive data. In the health and physiological sensing domain, federated learning has significant potential. Specifically in our scenario, FL means that facial video data and physiological gold-standard signals can remain on the mobile device and/or be processed in real-time and not transferred to any cloud storage. By only updating model parameters to the centralized server, we can learn a shared model through aggregating a large diverse population without collecting their own data.  

As a baseline, we use FedAvg \cite{mcmahan2017communication}, the most commonly used federated learning algorithm. As Fig.~\ref{fig:overview} illustrates, each client uses video recordings and reference PPG signals captured by the owner of the device. These are used to train models local to each client. The model weights are then uploaded to a centralized server to execute model aggregation. 
FedAvg~\cite{mcmahan2017communication} uses an iterative model averaging approach to updating the model server's model's weights. This approach has been shown to be effective on image classification tasks so we start with this technique as a baseline for creating camera-based physiological measurement models in a federate manner.

\begin{figure*}[t!]
  \includegraphics[width=\textwidth]{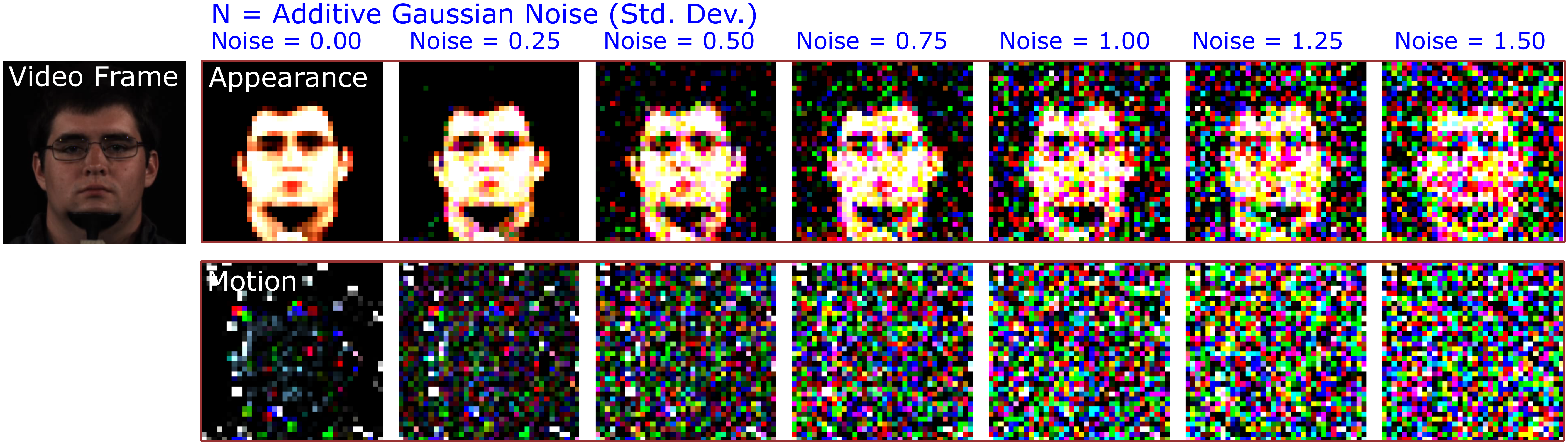}
  \caption{In our experiments we simulate camera sensor noise by adding Gaussian noise to the images. Here we illustrate the impact on the appearance and motion inputs to the two branch convolutional attention network.}
  \label{fig:vis_noise}
\end{figure*}

\begin{figure*}[t!]
  \includegraphics[width=\textwidth]{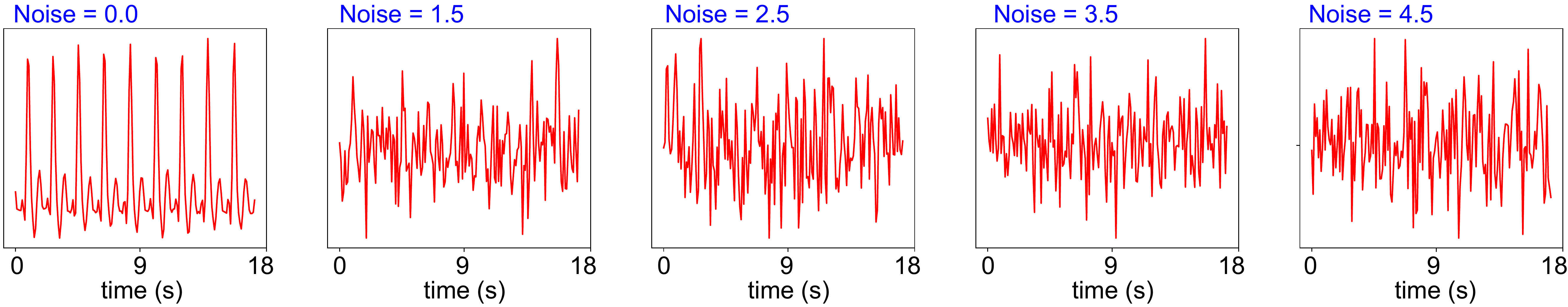}
  \caption{In our experiments we simulate contact reference PPG sensor noise by adding Gaussian noise to gold-standard contact sensor measurements. Here we illustrate the impact on training labels.}
  \label{fig:vis_label_noise}
\end{figure*}

\textbf{Noise Weighted Federated Learning.}
When training video-based physiological measurement algorithms, the goal is to recover physiological changes from very subtle (often sub-pixel) variations in image intensity. As we shall see training with FedAvg is effective if the training data from every client is ``clean'' (i.e., not corrupted). However, in reality it is much more likely to be the case that the quality of the training data on some individual devices will be better than others. This could be due to camera noise (e.g., quantization error) which can be most severe in poor lighting conditions when the gain is increased or user error in collecting and synchronizing the videos and reference physiological signals.

Treating the weights from every client equally is naive and does not appear to be the best way to solve optimization if the quality of the data from some devices is worse than that from others. 
We would prefer to have a method that promotes weights from clients with less noisy data (exploitation) while still considering weights from all clients to promote diversity (exploration). 
In this paper, we propose a simple but effective version of federated averaging, called \projectname, by leveraging knowledge about the signal quality from each client. The centralized server model weight is calculated as in Equation \ref{eq:signal_quality} where $k$ is the index of a layer, $\sigma_{i}$ is the signal quality of client $i$, $\omega^k_{i}$ is the client $i$'s model weights in the layer $k$ , $b^k_{i}$ is the bias in the client model weights in the layer $k$. 

\begin{equation} \label{eq:signal_quality}
     W^k_{server} = \frac{\sigma_i}{\sum{\sigma_{i}}} \cdot (\omega^k_{i} + b^k_{i})
\end{equation}

Our proposed signal-based aggregation is outlined in Algorithm 1. We first have an initialized centralized model weight $W_0$. Within each round of federated training, we randomly select a subset of clients for training. For each selected client, we then run a one-step optimization. After finishing local training for all the selected clients, we then perform signal-quality based aggregation as Equation \ref{eq:signal_quality} does. The output of each round in federated training is an aggregated model based on signal quality of selected clients' weights. Unlike FedAvg, which treats weights from all clients equally during model aggregation, our proposed leverages the fact that signal quality has a big impact on model performance to perform a more adaptive form of aggregation.

\section{Experiment}

\subsection{Datasets}
\textbf{AFRL} \citep{estepp2014recovering}: There is a total of 300 videos from 17 male participants and 8 female participants. The resolution of each video is 658 x 492 and the sampling rate is 120 fps. We down-sampled resolution to 36 x 36 \cite{chen2018deepphys} and resampled the video to 30 fps. A fingertip reflectance medical-grade photoplethysmograms (PPG) device was provided to record ground-truth PPG signal for training the network and for evaluating the performance of our proposed system. During the data collection, every participant was asked to keep stationary for the first two tasks and perform head motion tasks in the subsequent four tasks. These motion tasks include rotating their head along the vertical axis, horizontal axis as well as orienting their head randomly to one of nine predefined locations. For the vertical and horizontal rotations, participants were asked to rotate in an angular velocity of 10 degrees/second, 20 degrees/second, 30 degrees/second, respectively. The six recording were repeated twice with two backgrounds. This data collection protocol was approved by the institutions IRB.

\textbf{MMSE-HR} \citep{zhang2016multimodal}: 40 participants were recruited to join the data collection, and there is a total of 102 videos at resolution of 1040 x 1392 and sampling rate of 25 fps. The ground-truth PPG signal was recorded by a Biopac2 MP150 system\footnote{\url{https://www.biopac.com/}} at 1000 fps. These size of this dataset is smaller than AFRL, but it include more spontaneous motions videos such as emotions. This data collection protocol was approved by the institutions IRB. 

\textbf{UBFC} \citep{bobbia2019unsupervised}: A total of 42 videos from 42 participants were recorded at resolution of 640 x 480 and sampling rate of 30 fps. UBFC has a similar volume as MMSE, which is also smaller than AFRL. All the videos are recorded at uncompressed 8-bit RGB format. The medical-grade pulse oximeter (CMS50E transmissive pulse oximeter) was used to record PPG signal for evaluation. All the participants were asked to keep stationary during the experiments. This data collection protocol was approved by the institutions IRB.

\subsection{Implementation Details} 

We implemented our system in PyTorch \citep{paszke2019pytorch}, and all the experiments were conducted on an Nvidia 2080Ti GPU. We chose TS-CAN \cite{liu2020multi} as our backbone network to evaluate how FL works in remote physiological measurement since TS-CAN is the state-of-the-art neural network and can process frames in real-time on mobile platforms. To briefly summarize, TS-CAN is a two-branch neural network for on-device camera-based physiological measurement. The network contains an appearance branch that takes a sequence of normalized frames as inputs and generates attention masks to guide TS-CAN's motion branch. The motion branch takes a sequence of normalized difference frames (difference between every two consecutive frames). TS-CAN also leverages tensor shift modules to efficiently model temporal relationships which helps extract the subtle physiological signals in the videos. More details can be found in~\cite{liu2020multi}. 

We first implemented TS-CAN with a window size of 20 frames instead of 10 frames because prior work has empirically shown a larger window size leads to better overall performance \cite{liu2021metaphys}. In this work, we focus on cross-dataset evaluation since the performance on cross-dataset evaluation is substantially worse than within-dataset evaluation using current state-of-the-art methods \citep{chen2018deepphys,liu2020multi}. We conducted all the federated training on the AFRL dataset~\cite{estepp2014recovering} and evaluated the aggregated model on UBFC~\cite{bobbia2019unsupervised} and MMSE~\cite{zhang2016multimodal} datasets. For the federated training, we chose the Adam optimizer \citep{kingma2014adam} with an learning rate of 0.001 on the client updates. We trained all the federated experiments for seven rounds until convergence. We followed the same training schema to replicate the traditional supervised performance of TS-CAN \cite{liu2020multi, liu2021metaphys}.  

To simulate different levels of noise in our training data (AFRL), we first sampled a subject noise level, $\sigma_s$, for each of the 25 subjects in the dataset from a Gaussian distribution with a mean equal to the experiment noise level (e.g. 0.25) and standard deviation of 0.1.
During the training, to add noise to the videos we added Gaussian pixel noise from another distribution with mean of zero and standard deviation at the subject's noise level, $\sigma_s$. To add noise to the labels we added a vector of Gaussian noise from a distribution with mean of zero and standard deviation at the subject's noise level, $\sigma_s$. 
These noise samples were then were added to each video frames or ground-truth label vector, respectively, as the Fig. \ref{fig:vis_noise} and \ref{fig:vis_label_noise} illustrate. In the federated weighting process, the signal quality score was assigned to $\sigma_s$ after normalizing across all subjects. As Fig. \ref{fig:vis_noise} and \ref{fig:vis_label_noise} show, we performed experiments adding six levels of noise to the videos [0.25, 0.50, 0.75, 1.00, 1.25, 1.50], and four levels of noise to the ground-truth labels [1.5, 2.5, 3.5, 4.5], respectively. 

Since our network is trained on the derivative of the PPG signal~\cite{chen2018deepphys}. We applied standard post-processing steps to extract the heart rate estimate: 1) calculating cumulative sum and using a detrending function~\cite{tarvainen2002advanced} ($\lambda$=10) to convert the signal to the PPG waveform; 2) dividing the estimated and ground-truth values for each participant into 360-frame non-overlapping moving windows (approximately 12 seconds); 3) applying a 2nd-order Butterworth filter with a cutoff frequency of 0.75 and 2.5 Hz which represents a realistic range of heart rates for adults. Following those steps, we then computed three metrics for each window including the mean absolute error (MAE) in heart rate frequency between the predicted signal and the reference contact PPG, signal-to-noise ratio (SNR)~\cite{de2013robust} of the waveform and the Pearson correlation coefficient between the heart rate estimates and the those from the reference contact PPG. For heart rate estimation the frequency of the heart rate was determined by selecting the frequency with maximum power in the range [40Hz, 150Hz]. 

To explore the efficiency of end-to-end deployment in on-device training and inference, we also conducted experiments on a quad-core Cortex-A72 Raspberry Pi 4B to evaluate the model's performance on an edge device. We trained the model and performed inference 10 times to get a reliable averaged on-device training and inference time.

\section{Results \& Discussion}
\label{sec:dis_results}

\begin{table*}
	\caption{Comparison between traditional supervised training and FL with noise level of 0. Bold numbers reflect better performance.}
	\vspace{0.3cm}
	\label{tab:fl_comparsion}
	\centering
	\small
	\setlength\tabcolsep{3pt} 
	\begin{tabular}{r|ccc|ccc}
	\toprule
		& \multicolumn{3}{c}{\textbf{UBFC}} & \multicolumn{3}{c}{\textbf{MMSE}} \\
        \textbf{Method} & MAE$\downarrow$ &  SNR$\uparrow$ & Pearson$\uparrow$ & MAE$\downarrow$ &  SNR$\uparrow$ & Pearson$\uparrow$ \\ \hline  \hline 
        Supervised Training \cite{liu2020multi} & 2.31 & 4.34 & 0.93 & \textbf{2.99} & \textbf{2.42} & \textbf{0.79}  \\
        Federated Training & \textbf{2.00} & \textbf{4.38} & 0.93 & 3.65 &  1.45 & 0.77 \\ 
    \bottomrule 
   \end{tabular}\\
      \tiny
  MAE = Mean Absolute Error in HR estimation, SNR = BVP Signal-to-Noise Ratio, $\rho$ = Pearson Correlation in HR estimation.
    \vspace{-0.1cm}
\end{table*}

\begin{table*}
\caption{Comparison between FedAvg and \projectname with different levels of video noise.}
\label{tab:video_noisy}
\small
\centering
\begin{tabular}{lr|cc|cc|cc}
\toprule
 & \multicolumn{1}{l}{} & \multicolumn{2}{c|}{\textbf{MAE (beats/min)}$\downarrow$} & \multicolumn{2}{c|}{\textbf{SNR (dB)}$\uparrow$} & \multicolumn{2}{c}{\textbf{Pearson}$\uparrow$} \\
\multicolumn{1}{c}{\textbf{Dataset}} & \textbf{Noise} & \textbf{FedAvg} & \textbf{\projectname} & \textbf{FedAvg} & \textbf{\projectname} & \textbf{FedAvg} & \textbf{\projectname} \\ \hline \hline
 & 0 & 2.00 & 2.00 & 4.38 & 4.38 & 0.93 & 0.93 \\ 
& 0.25 & 3.06 & \textbf{2.44} & 2.33 & \textbf{3.53} & 0.83 & \textbf{0.92} \\ 
 & 0.50 & 4.14 & \textbf{2.90} & 1.69 & \textbf{2.01} & 0.76 & \textbf{0.89} \\ 
\textbf{UBFC} & 0.75 & 4.59 & \textbf{3.47} & 0.02 & \textbf{2.07} & 0.76 & \textbf{0.87} \\ 
 & 1.00 & 5.18 & \textbf{4.16} & -1.18 & \textbf{0.4} & 0.75 & \textbf{0.81} \\ 
 & 1.25 & 7.48 & \textbf{7.02} & \textbf{-2.77} & -3.22 & 0.66 & \textbf{0.79} \\ 
 & 1.50 & 7.44 & \textbf{4.59} & -2.33 & \textbf{-0.03} & 0.66 & \textbf{0.79} \\ \hline
 & 0 & 3.93 & 3.93 & 2.29 & 2.29 & 0.80 & 0.80 \\ 
& 0.25 & 4.58 & \textbf{4.33} & 0.67 & \textbf{0.84} & 0.65 & \textbf{0.67} \\ 
 & 0.50 & 5.22 & \textbf{4.44} & 0.07 & \textbf{0.41} & 0.57 & \textbf{0.68} \\ 
\textbf{MMSE} & 0.75 & 6.46 & \textbf{5.38} & -0.51 & \textbf{0.07} & 0.46 & \textbf{0.54} \\ 
 & 1.00 & 6.58 & \textbf{5.39} & -1.12 & \textbf{0.01} & 0.44 & \textbf{0.56} \\ 
 & 1.25 & 6.61 & \textbf{5.77} & -0.90 & \textbf{-0.64} & 0.44 & \textbf{0.53} \\ 
 & 1.50 & 6.92 & \textbf{6.17} & -2.29 & \textbf{-1.79} & 0.43 & \textbf{0.55} \\ 
 \bottomrule 
\end{tabular}\\
      \tiny
   MAE = Mean Absolute Error in HR estimation, SNR = BVP Signal-to-Noise Ratio, $\rho$ = Pearson Correlation in HR estimation.
\end{table*}

\begin{table*}[!h]
\caption{Comparison between FedAvg and \projectname with different levels of label noise.}
\label{tab:label_noisy}
\small
\centering
\begin{tabular}{lr|cc|cc|cc}
\toprule 
 & \multicolumn{1}{l|}{} & \multicolumn{2}{c|}{\textbf{MAE (beats/min)}$\downarrow$} & \multicolumn{2}{c|}{\textbf{SNR (dB)}$\uparrow$} & \multicolumn{2}{c}{\textbf{Pearson}$\uparrow$} \\ 
\multicolumn{1}{c}{\textbf{Dataset}} & \textbf{Noise} & \textbf{FedAvg} & \textbf{\projectname} & \textbf{FedAvg} & \textbf{\projectname} & \textbf{FedAvg} & \textbf{\projectname} \\ \hline \hline
 & 0 & 2.00 & 2.00 & 4.38 & 4.38 & 0.93 & 0.93 \\ 
 & 1.5 & \textbf{1.79} & 2.41 &  \textbf{4.72} &  4.70 &   \textbf{0.96} & 0.93 \\ 
\textbf{UBFC} & 2.5 & 2.05 & \textbf{2.02} & \textbf{4.83} & 4.69 &  0.94 & \textbf{0.96} \\ 
 & 3.5 & \textbf{1.88} & 2.67 & \textbf{4.28} & 3.44 & \textbf{0.96} & 0.93 \\ 
 & 4.5 & 2.73 & \textbf{2.16} & 3.94 & \textbf{4.97} & \textbf{0.96} & 0.94 \\ \hline
 & 0 & 3.93 & 3.93 & 2.29 & 2.29 & 0.80 & 0.80 \\ 
 & 1.5 & \textbf{3.72} & 4.07 &  \textbf{0.97} & -0.27 &  \textbf{0.78} &  0.73 \\ 
\textbf{MMSE} & 2.5 & 4.60 & \textbf{3.88} & -0.28 &  \textbf{0.36} &  0.73 & \textbf{0.79} \\ 
 & 3.5 & 4.44 & \textbf{3.91} & -0.34 & \textbf{0.38} &  0.74 & \textbf{0.79} \\ 
 & 4.5 & 4.94 & \textbf{4.42} & -0.81 & \textbf{-0.78} & 0.72 &  \textbf{0.74} \\ 
 \bottomrule 
\end{tabular}\\
      \tiny
   MAE = Mean Absolute Error in HR estimation, SNR = BVP Signal-to-Noise Ratio, $\rho$ = Pearson Correlation in HR estimation.
\end{table*}

\textbf{How does FL compare to regular supervised training?} The results of regular supervised training and FedAvg FL are summarized in Table \ref{tab:fl_comparsion}. For the UBFC dataset, FL outperforms regular supervised training. On the other hand, regular supervised training outperforms FL on the MMSE dataset. Through this comparison, we observe that the differences are small and that there is not a consistent accuracy difference between the two. However, FL has several additional benefits compared to regular training as have been discussed. Therefore, our results point to a promising future for FL in privacy preserving camera-based cardiac measurement. 

\begin{figure}[t]
  \centering
  \includegraphics[width=1.0\columnwidth]{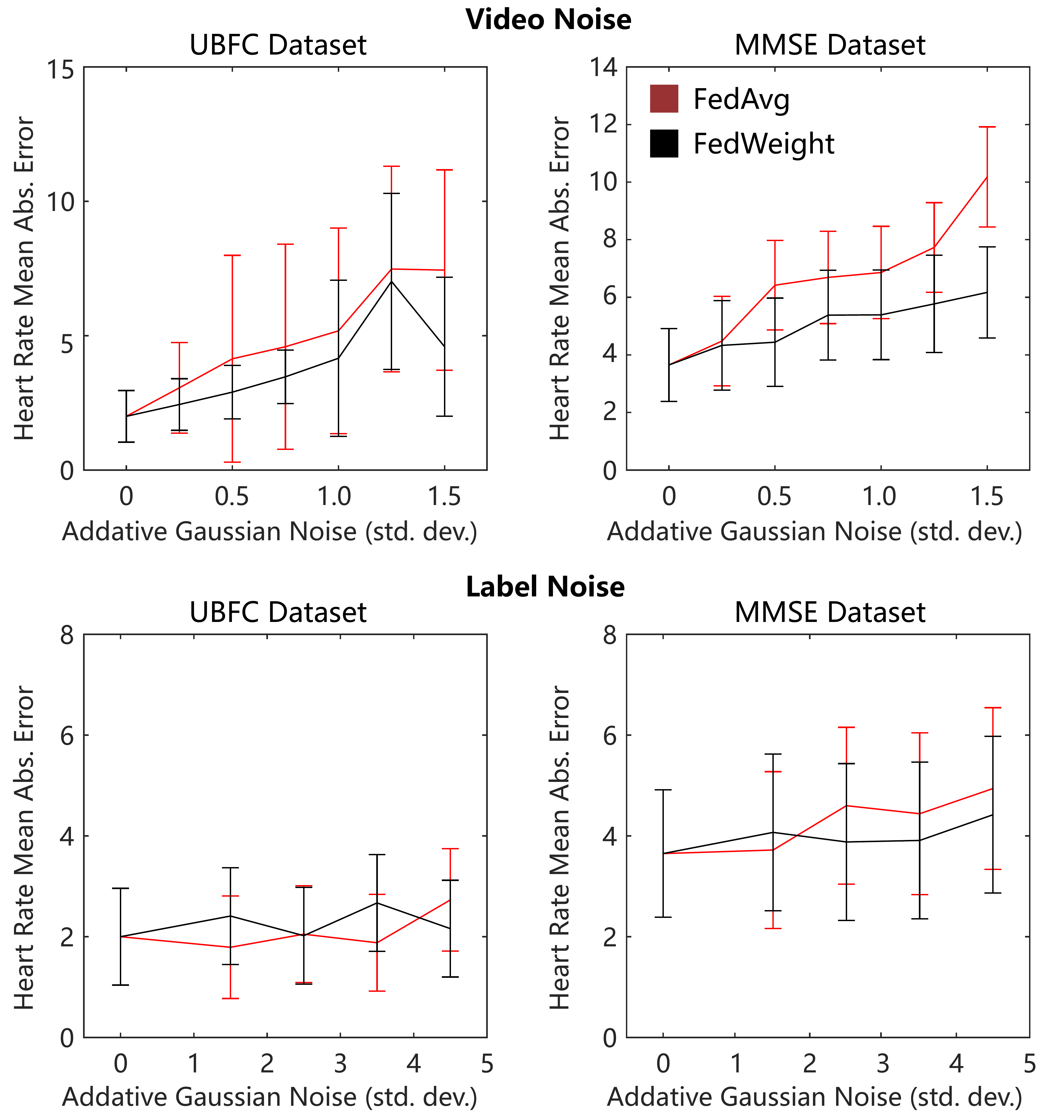}
  \caption{The heart rate mean absolute error for FedAvg and \projectname at different video/label noise levels in the UBFC and MMSE datasets. Error bars reflect standard error where N is the number of videos.}
  \label{fig:noiseresults}
    \vspace{-0.2cm}
\end{figure}

\textbf{How does video and label noise impact FL?}
Next, we examine how the performance of FL is affected by noise in the videos and labels. Tables \ref{tab:video_noisy} and \ref{tab:label_noisy} and Fig.~\ref{fig:noiseresults} show that the performance of the camera-based pulse measurement and heart rate estimation degrades significantly when using a naive weight averaging when some of the data is corrupted by noise. For example, in the noisy video experiments, we observed that the HR MAE increases by 19\% and 20\% when the noise level was increased from 0.25 to 0.5 and from 0.5 to 0.75 (UBFC dataset). 
However, a different pattern was found in the noisy label experiments described in Table \ref{tab:label_noisy}. The MAE results remain similar across different noise levels, which indicates that noisy label does not significantly affect the performance of training and could be used as a regularization technique during training. Overall, the label noise had a much less severe impact on performance. In summary, simple federated averaging struggles with either noisy data or noisy labels in remote physiological measurement.

\textbf{What is the impact of \projectname?}
For the video noise level of 0.25, 0.5, 0.75, 1.0, 1.25 and 1.5, \projectname improves 20\%, 30\%, 24\%, 20\%, 6\%  and 38\% in MAE respectively, when compared to FedAvg. A similar pattern was also observed in the MMSE dataset where \projectname leads to a reduction of errors by 5\%, 15\%, 17\%, 18\%, 13\% and 11\% respectively. Moreover, our proposed \projectname achieved comparable results as FedAvg in the case of noisy labels on the UBFC dataset. \projectname helped achieve slightly better results in the MMSE dataset, but we still argue that noisy labels don't  significant affect the performance of federated training or traditional supervised training. To summarize, intelligently combining weights using a signal quality weighted averaging method leads to a considerably more robust model if the features (videos) are corrupted by noise. We believe that this result would likely by consistent for many other computer vision and machine learning tasks.

\textbf{How to automate signal quality measurement? }
In this paper, we assume the noise level and signal quality are available to the centralized server. This could be the case if clients were able to provide a data quality report based on their knowledge of their individual sensor noise profiles. However, automating signal quality measurement would be preferred in many real-world scenarios. We are aware of this limitation and actively working on building an range of automatic signal quality metrics to test. Inspired by the metric in the task of super resolution, we argue that Peak Signal-to-Noise Ratio (PSNR) could be one way of measuring image noise level and quality. Moreover, we are also actively studying using the patterns of training loss and the quality of estimated PPG signal to assess the quality of videos.

\textbf{Can we create an on-device FL prototype?} 

We deployed our FL system on-device as part of our experimentation.
The average on-device inference time was 24.5ms per frame while the on-device training time was 105ms per frame. Based on these results, the training time is almost five time the inference time. 
Deploying models like our on edge devices is non-trivial. Most deep learning frameworks \cite{tensorflow, pytorch, mxnet} focus on training on server machines, leaving inference to edge devices \cite{tflite, efficient-inference}.  To enable efficient federated learning on edge devices, several challenges need to be solved: the underlying framework needs to allow efficient local training on the heterogeneous device; the runtime has to be small enough to fit on to a resource-constrained device; flexible communication patterns should be supported and simple to implement for different aggregation algorithms. We are actively exploring this direction based on deep learning compilation techniques \cite{tvm, glow}, including extending current deep learning compilers to training workload and optimize kernels for heterogeneous devices automatically.

\section{Limitations}

Although our proposed \projectname improves on the performance of federated camera-based physiological measurement in the presence of noise, there are still a few limitations. First, we picked six representative video noise levels and four label noise levels. However, these noise levels do not represent the entire spectrum of real-world noise. We plan to run greedy search experiments to explore more noise levels in the future. Second, we assume the ``ground-truth'' noise levels are available to the centralized server during model aggregation. In the future, we plan to develop a system to automatically measure noise levels and signal quality using domain knowledge (e.g., skewness of PPG signal and PSNR in the image) in imaging and physiology as discussed in section \ref{sec:dis_results}. Finally, we performance experiments on datasets that are not fully representative of all physical appearances. Before similar sensing algorithms are deployed they would require further validation and clinical evaluation. 
\section{Broader Impact}

Ubiquitous computing offers a lot of potential for improving access to healthcare. For those that find it difficult to, or cannot, travel to a physician easily would benefit from technology that provides reliable measurement of physiological vital signs. If measurement can be performed from only a video, what happens if we detect a health condition in an individual when analyzing a video for other purposes. When and how should that information be disclosed? If the system fails in a context where a person is in a remote location, it may lead them to panic. For example, non-contact camera-based vital sensing can be used to measure a person's stress level without any notification. Especially during this pandemic, video conference meeting has become the major way to communicate between people. Non-contact physiological sensing could be easily plugged in softwares such as Zoom or Teams. Employer could easily sense their employees' health status during the meeting if we don't have the law enforcement for th is technology. 

In the United States, a high standard was set by the Health Insurance Portability and Accountability Act (HIPAA) to protect sensitive patient data. We believe non-contat camera-based physiological measurement also should be under HIPPA compliance. Given the unique characteristic of camera-based physiological measurement, it even includes more sensitive information (e.g., long facial videos) than many other healthcare technology. We argue that a special protection of data transferring should be enforced to minimizing the risk of data leaking. A better way to do this is to store and run inference on local mobile devices. However, how to collect large-scale physiological and video data to train a "super" model still remains challenge due to the concerns of data leaking and management. In this paper, we have successfully demonstrated how federated learning interplays with non-contact physiological sensing. Even without uploading a single raw video or physiological data to centralized server, it is still possible to attain a "super" aggregated model for everyone to use.

\section{Conclusion}

In this paper, We present a federated learning system called \projectname that accounts for training imperfect data such as noisy data or noisy labels. We apply this to the task of camera-based remote physiological measurement. Our results show that traditional federated weight averaging degrades quickly if the data on some of the clients is corrupted by noise, our proposed method is more robust to corruption particularly video noise.  Federated learning has many attractive properties for camera-based health monitoring where it not only protect sensitive information but also provides a way to aggregate large scale clients to train a robust model. We envision federated learning and \projectname will have a big potential in various applications in mobile health, especially in remote physiological measurement.

{\small
\bibliographystyle{ieee_fullname}
\bibliography{egbib}

\begin{thebibliography}{10}\itemsep=-1pt

\bibitem{tensorflow}
Mart\'{\i}n Abadi, Ashish Agarwal, Paul Barham, Eugene Brevdo, Zhifeng Chen,
  Craig Citro, Greg~S. Corrado, Andy Davis, Jeffrey Dean, Matthieu Devin,
  Sanjay Ghemawat, Ian Goodfellow, Andrew Harp, Geoffrey Irving, Michael Isard,
  Yangqing Jia, Rafal Jozefowicz, Lukasz Kaiser, Manjunath Kudlur, Josh
  Levenberg, Dandelion Man\'{e}, Rajat Monga, Sherry Moore, Derek Murray, Chris
  Olah, Mike Schuster, Jonathon Shlens, Benoit Steiner, Ilya Sutskever, Kunal
  Talwar, Paul Tucker, Vincent Vanhoucke, Vijay Vasudevan, Fernanda Vi\'{e}gas,
  Oriol Vinyals, Pete Warden, Martin Wattenberg, Martin Wicke, Yuan Yu, and
  Xiaoqiang Zheng.
\newblock {TensorFlow}: Large-scale machine learning on heterogeneous systems,
  2015.
\newblock Software available from tensorflow.org.

\bibitem{bobbia2019unsupervised}
Serge Bobbia, Richard Macwan, Yannick Benezeth, Alamin Mansouri, and Julien
  Dubois.
\newblock Unsupervised skin tissue segmentation for remote
  photoplethysmography.
\newblock {\em Pattern Recognition Letters}, 124:82--90, 2019.

\bibitem{brisimi2018federated}
Theodora~S Brisimi, Ruidi Chen, Theofanie Mela, Alex Olshevsky, Ioannis~Ch
  Paschalidis, and Wei Shi.
\newblock Federated learning of predictive models from federated electronic
  health records.
\newblock {\em International journal of medical informatics}, 112:59--67, 2018.

\bibitem{brophy2021estimation}
Eoin Brophy, Maarten De~Vos, Geraldine Boylan, and Tomas Ward.
\newblock Estimation of continuous blood pressure from ppg via a federated
  learning approach.
\newblock {\em arXiv preprint arXiv:2102.12245}, 2021.

\bibitem{chan2016diagnostic}
Pak-Hei Chan, Chun-Ka Wong, Yukkee~C Poh, Louise Pun, Wangie Wan-Chiu Leung,
  Yu-Fai Wong, Michelle Man-Ying Wong, Ming-Zher Poh, Daniel Wai-Sing Chu, and
  Chung-Wah Siu.
\newblock Diagnostic performance of a smartphone-based photoplethysmographic
  application for atrial fibrillation screening in a primary care setting.
\newblock {\em Journal of the American Heart Association}, 5(7):e003428, 2016.

\bibitem{mxnet}
Tianqi Chen, Mu Li, Yutian Li, Min Lin, Naiyan Wang, Minjie Wang, Tianjun Xiao,
  Bing Xu, Chiyuan Zhang, and Zheng Zhang.
\newblock Mxnet: A flexible and efficient machine learning library for
  heterogeneous distributed systems, 2015.

\bibitem{tvm}
Tianqi Chen, Thierry Moreau, Ziheng Jiang, Lianmin Zheng, Eddie Yan, Haichen
  Shen, Meghan Cowan, Leyuan Wang, Yuwei Hu, Luis Ceze, et~al.
\newblock $\{$TVM$\}$: An automated end-to-end optimizing compiler for deep
  learning.
\newblock In {\em 13th $\{$USENIX$\}$ Symposium on Operating Systems Design and
  Implementation ($\{$OSDI$\}$ 18)}, pages 578--594, 2018.

\bibitem{chen2018deepphys}
Weixuan Chen and Daniel McDuff.
\newblock Deepphys: Video-based physiological measurement using convolutional
  attention networks.
\newblock In {\em Proceedings of the European Conference on Computer Vision
  (ECCV)}, pages 349--365, 2018.

\bibitem{chen2020fedhealth}
Yiqiang Chen, Xin Qin, Jindong Wang, Chaohui Yu, and Wen Gao.
\newblock Fedhealth: A federated transfer learning framework for wearable
  healthcare.
\newblock {\em IEEE Intelligent Systems}, 35(4):83--93, 2020.

\bibitem{de2013robust}
Gerard De~Haan and Vincent Jeanne.
\newblock Robust pulse rate from chrominance-based rppg.
\newblock {\em IEEE Transactions on Biomedical Engineering}, 60(10):2878--2886,
  2013.

\bibitem{estepp2014recovering}
Justin~R Estepp, Ethan~B Blackford, and Christopher~M Meier.
\newblock Recovering pulse rate during motion artifact with a multi-imager
  array for non-contact imaging photoplethysmography.
\newblock In {\em 2014 IEEE International Conference on Systems, Man, and
  Cybernetics (SMC)}, pages 1462--1469. IEEE, 2014.

\bibitem{hernandez2015bioinsights}
Javier Hernandez, Daniel~J McDuff, and Rosalind~W Picard.
\newblock Bioinsights: Extracting personal data from “still” wearable
  motion sensors.
\newblock In {\em 2015 IEEE 12th International Conference on Wearable and
  Implantable Body Sensor Networks (BSN)}, pages 1--6. IEEE, 2015.

\bibitem{efficient-inference}
Ziheng Jiang, Tianqi Chen, and Mu Li.
\newblock Efficient deep learning inference on edge devices.
\newblock {\em ACM SysML}, 2018.

\bibitem{kingma2014adam}
Diederik~P Kingma and Jimmy Ba.
\newblock Adam: A method for stochastic optimization.
\newblock {\em arXiv preprint arXiv:1412.6980}, 2014.

\bibitem{konevcny2016federated}
Jakub Kone{\v{c}}n{\`y}, H~Brendan McMahan, Felix~X Yu, Peter Richt{\'a}rik,
  Ananda~Theertha Suresh, and Dave Bacon.
\newblock Federated learning: Strategies for improving communication
  efficiency.
\newblock {\em arXiv preprint arXiv:1610.05492}, 2016.

\bibitem{lee2020meta}
Eugene Lee, Evan Chen, and Chen-Yi Lee.
\newblock Meta-rppg: Remote heart rate estimation using a transductive
  meta-learner.
\newblock {\em Proceedings of the European Conference on Computer Vision
  (ECCV)}, 2020.

\bibitem{tflite}
Juhyun Lee, Nikolay Chirkov, Ekaterina Ignasheva, Yury Pisarchyk, Mogan Shieh,
  Fabio Riccardi, Raman Sarokin, Andrei Kulik, and Matthias Grundmann.
\newblock On-device neural net inference with mobile gpus, 2019.

\bibitem{li2014remote}
Xiaobai Li, Jie Chen, Guoying Zhao, and Matti Pietikainen.
\newblock Remote heart rate measurement from face videos under realistic
  situations.
\newblock In {\em Proceedings of the IEEE conference on computer vision and
  pattern recognition}, pages 4264--4271, 2014.

\bibitem{liu2020multi}
Xin Liu, Josh Fromm, Shwetak Patel, and Daniel McDuff.
\newblock Multi-task temporal shift attention networks for on-device
  contactless vitals measurement.
\newblock {\em arXiv preprint arXiv:2006.03790}, 2020.

\bibitem{liu2021metaphys}
Xin Liu, Ziheng Jiang, Josh Fromm, Xuhai Xu, Shwetak Patel, and Daniel McDuff.
\newblock Metaphys: Few-shot adaptation for non-contact physiological
  measurement.
\newblock In {\em Proceedings of the Conference on Health, Inference, and
  Learning}, pages 154--163, 2021.

\bibitem{mcduff2021camera}
Daniel McDuff.
\newblock Camera measurement of physiological vital signs.
\newblock {\em arXiv preprint arXiv:2111.11547}, 2021.

\bibitem{mcduff2017impact}
Daniel~J McDuff, Ethan~B Blackford, and Justin~R Estepp.
\newblock The impact of video compression on remote cardiac pulse measurement
  using imaging photoplethysmography.
\newblock In {\em 2017 12th IEEE International Conference on Automatic Face \&
  Gesture Recognition (FG 2017)}, pages 63--70. IEEE, 2017.

\bibitem{mcmahan2017communication}
Brendan McMahan, Eider Moore, Daniel Ramage, Seth Hampson, and Blaise~Aguera y
  Arcas.
\newblock Communication-efficient learning of deep networks from decentralized
  data.
\newblock In {\em Artificial Intelligence and Statistics}, pages 1273--1282.
  PMLR, 2017.

\bibitem{pytorch}
Adam Paszke, Sam Gross, Francisco Massa, Adam Lerer, James Bradbury, Gregory
  Chanan, Trevor Killeen, Zeming Lin, Natalia Gimelshein, Luca Antiga, Alban
  Desmaison, Andreas Kopf, Edward Yang, Zachary DeVito, Martin Raison, Alykhan
  Tejani, Sasank Chilamkurthy, Benoit Steiner, Lu Fang, Junjie Bai, and Soumith
  Chintala.
\newblock Pytorch: An imperative style, high-performance deep learning library.
\newblock In H. Wallach, H. Larochelle, A. Beygelzimer, F. d\textquotesingle
  Alch\'{e}-Buc, E. Fox, and R. Garnett, editors, {\em Advances in Neural
  Information Processing Systems 32}, pages 8024--8035. Curran Associates,
  Inc., 2019.

\bibitem{paszke2019pytorch}
Adam Paszke, Sam Gross, Francisco Massa, Adam Lerer, James Bradbury, Gregory
  Chanan, Trevor Killeen, Zeming Lin, Natalia Gimelshein, Luca Antiga, et~al.
\newblock Pytorch: An imperative style, high-performance deep learning library.
\newblock In {\em Advances in neural information processing systems}, pages
  8026--8037, 2019.

\bibitem{poh2010advancements}
Ming-Zher Poh, Daniel~J McDuff, and Rosalind~W Picard.
\newblock Advancements in noncontact, multiparameter physiological measurements
  using a webcam.
\newblock {\em IEEE transactions on biomedical engineering}, 58(1):7--11, 2010.

\bibitem{poh2010non}
Ming-Zher Poh, Daniel~J McDuff, and Rosalind~W Picard.
\newblock Non-contact, automated cardiac pulse measurements using video imaging
  and blind source separation.
\newblock {\em Optics express}, 18(10):10762--10774, 2010.

\bibitem{qayyum2021collaborative}
Adnan Qayyum, Kashif Ahmad, Muhammad~Ahtazaz Ahsan, Ala Al-Fuqaha, and Junaid
  Qadir.
\newblock Collaborative federated learning for healthcare: Multi-modal covid-19
  diagnosis at the edge.
\newblock {\em arXiv preprint arXiv:2101.07511}, 2021.

\bibitem{rieke2020future}
Nicola Rieke, Jonny Hancox, Wenqi Li, Fausto Milletari, Holger~R Roth, Shadi
  Albarqouni, Spyridon Bakas, Mathieu~N Galtier, Bennett~A Landman, Klaus
  Maier-Hein, et~al.
\newblock The future of digital health with federated learning.
\newblock {\em NPJ digital medicine}, 3(1):1--7, 2020.

\bibitem{glow}
Nadav Rotem, Jordan Fix, Saleem Abdulrasool, Garret Catron, Summer Deng, Roman
  Dzhabarov, Nick Gibson, James Hegeman, Meghan Lele, Roman Levenstein, Jack
  Montgomery, Bert Maher, Satish Nadathur, Jakob Olesen, Jongsoo Park, Artem
  Rakhov, Misha Smelyanskiy, and Man Wang.
\newblock Glow: Graph lowering compiler techniques for neural networks, 2019.

\bibitem{takano2007heart}
Chihiro Takano and Yuji Ohta.
\newblock Heart rate measurement based on a time-lapse image.
\newblock {\em Medical engineering \& physics}, 29(8):853--857, 2007.

\bibitem{tarassenko2014non}
L Tarassenko, M Villarroel, A Guazzi, J Jorge, DA Clifton, and C Pugh.
\newblock Non-contact video-based vital sign monitoring using ambient light and
  auto-regressive models.
\newblock {\em Physiological measurement}, 35(5):807, 2014.

\bibitem{tarvainen2002advanced}
Mika~P Tarvainen, Perttu~O Ranta-Aho, and Pasi~A Karjalainen.
\newblock An advanced detrending method with application to hrv analysis.
\newblock {\em IEEE Transactions on Biomedical Engineering}, 49(2):172--175,
  2002.

\bibitem{tulyakov2016self}
Sergey Tulyakov, Xavier Alameda-Pineda, Elisa Ricci, Lijun Yin, Jeffrey~F Cohn,
  and Nicu Sebe.
\newblock Self-adaptive matrix completion for heart rate estimation from face
  videos under realistic conditions.
\newblock In {\em Proceedings of the IEEE conference on computer vision and
  pattern recognition}, pages 2396--2404, 2016.

\bibitem{verkruysse2008remote}
Wim Verkruysse, Lars~O Svaasand, and J~Stuart Nelson.
\newblock Remote plethysmographic imaging using ambient light.
\newblock {\em Optics express}, 16(26):21434--21445, 2008.

\bibitem{wang2016algorithmic}
Wenjin Wang, Albertus~C den Brinker, Sander Stuijk, and Gerard de Haan.
\newblock Algorithmic principles of remote ppg.
\newblock {\em IEEE Transactions on Biomedical Engineering}, 64(7):1479--1491,
  2016.

\bibitem{xu2020pathological}
Zhe Xu, Lei Shi, Yijin Wang, Jiyuan Zhang, Lei Huang, Chao Zhang, Shuhong Liu,
  Peng Zhao, Hongxia Liu, Li Zhu, et~al.
\newblock Pathological findings of covid-19 associated with acute respiratory
  distress syndrome.
\newblock {\em The Lancet respiratory medicine}, 8(4):420--422, 2020.

\bibitem{yang2019federated}
Qiang Yang, Yang Liu, Tianjian Chen, and Yongxin Tong.
\newblock Federated machine learning: Concept and applications.
\newblock {\em ACM Transactions on Intelligent Systems and Technology (TIST)},
  10(2):1--19, 2019.

\bibitem{yu2019remote}
Zitong Yu, Xiaobai Li, and Guoying Zhao.
\newblock Remote photoplethysmograph signal measurement from facial videos
  using spatio-temporal networks.
\newblock In {\em Proc. BMVC}, pages 1--12, 2019.

\bibitem{zhang2016multimodal}
Zheng Zhang, Jeff~M Girard, Yue Wu, Xing Zhang, Peng Liu, Umur Ciftci, Shaun
  Canavan, Michael Reale, Andy Horowitz, Huiyuan Yang, et~al.
\newblock Multimodal spontaneous emotion corpus for human behavior analysis.
\newblock In {\em Proceedings of the IEEE Conference on Computer Vision and
  Pattern Recognition}, pages 3438--3446, 2016.

\bibitem{zheng2020covid}
Ying-Ying Zheng, Yi-Tong Ma, Jin-Ying Zhang, and Xiang Xie.
\newblock Covid-19 and the cardiovascular system.
\newblock {\em Nature Reviews Cardiology}, 17(5):259--260, 2020.

\end{thebibliography}
}

\end{document}